\documentclass[journal]{IEEEtran}

\ifCLASSINFOpdf

\else

\fi

\usepackage{graphicx}

\usepackage{amssymb}
 \usepackage{amsthm}
\usepackage{amsmath}
\usepackage{color}
\usepackage{setspace}
\usepackage{url}
\usepackage{algorithm}
\usepackage{algorithmic}
\usepackage{graphicx}
\usepackage{subfigure}
\usepackage{multirow}
\usepackage{array}
\usepackage[font=small,labelfont=bf,tableposition=top]{caption}
\usepackage{booktabs}
\usepackage{threeparttable}
\usepackage{stfloats}
\usepackage{color}
\renewcommand{\baselinestretch}{1.0}

\begin{document}


\title{LogoDet-3K: A Large-Scale Image Dataset for Logo Detection}

\author{Jing Wang, Weiqing Min, \textit{Member, IEEE}, Sujuan Hou, \textit{Member, IEEE}, Shengnan Ma, \\ Yuanjie Zheng, \textit{Member, IEEE}, Shuqiang Jiang, \textit{Senior Member, IEEE}
\thanks{J. Wang, S. Hou, S. Ma and Y. Zheng are School of Information Science and Engineering, Shandong Normal University, Shandong, 250358, China. Email: 2018020875@stu.sdnu.edu.cn, sujuanhou@sdnu.edu.cn, 201711030133@stu.sdnu.edu.cn,
zhengyuanjie@gmail.com. W. Min and S. Jiang are with the Key Laboratory of Intelligent Information Processing, Institute of Computing Technology, Chinese Academy of Sciences, Beijing, 100190, China, and also with University of Chinese Academy of Sciences, Beijing, 100049, China. Email: minweiqing@ict.ac.cn, sqjiang@ict.ac.cn.}
}
\markboth{IEEE Transactions on Multimedia,~Vol.~X,
No.~XX,~Month~Year}{}


\maketitle

\begin{abstract}
Logo detection has been gaining considerable attention because of its wide range of applications in the multimedia field, such as copyright infringement detection, brand visibility monitoring, and product brand management on social media. In this paper, we introduce LogoDet-3K, the largest logo detection dataset with full annotation, which has 3,000 logo categories, about 200,000 manually annotated logo objects and 158,652 images. LogoDet-3K creates a more challenging benchmark for logo detection, for its higher comprehensive coverage and wider variety in both logo categories and annotated objects compared with existing datasets. We describe the collection and annotation process of our dataset, analyze its scale and diversity in comparison to other datasets for logo detection. We further propose a strong baseline method Logo-Yolo, which incorporates Focal loss and CIoU loss into the state-of-the-art YOLOv3 framework for large-scale logo detection. Logo-Yolo can solve the problems of multi-scale objects, logo sample imbalance and inconsistent bounding-box regression. It obtains about 4\% improvement on the average performance compared with YOLOv3, and greater improvements compared with reported several deep detection models on LogoDet-3K. The evaluations on other three existing datasets further verify the effectiveness of our method, and demonstrate better generalization ability of LogoDet-3K on logo detection and retrieval tasks. The LogoDet-3K dataset is used to promote large-scale logo-related research and it can be found at {\color{red}{\url{https://github.com/Wangjing1551/LogoDet-3K-Dataset}}}.

\end{abstract}


\IEEEpeerreviewmaketitle

\section{Introduction}
Logo-related research has always been extensively studied in the field of multimedia~\cite{Brand14, Romberg2011Scalable, Revaud2012Correlation, Kalantidis2011STL, Romberg2013Bundle}. As an important branch of logo research, logo detection~\cite{Automatic2005Yan, Region2016, Eggert2017Improving} plays a critical role for its various applications and services, such as intelligent transportation~\cite{Yang2015Car}, brand visibility monitoring~\cite{Filtering2016gao} and analysis~\cite{Visual18}, trademark infringement detection~\cite{Brand14} and video advertising research~\cite{Video2017cheng}.

Currently, deep-learning approaches have been widely used in logo detection, like Faster R-CNN~\cite{Ren2015}, SSD~\cite{liu2016ssd} and YOLOv3~\cite{Joseph2018Yolov3}. By supporting the learning process of deep networks with millions of parameters, large-scale logo  datasets are crucial in logo detection. However, most existing logo researches focus on small-scale datasets, such as BelgaLogos~\cite{Neumann2001Integration} and FlickrLogos-32~\cite{Romberg2011Scalable}. Recently, although some large-scale logo datasets are proposed for recognition and detection, such as WebLogo-2M~\cite{Su2017WebLogo}, PL2K~\cite{II2019Scalable} and Logo-2K+~\cite{Wang2020Logo2K}, these logo datasets are  either only labeled on image-level~\cite{Su2017WebLogo, Wang2020Logo2K} or not publicly available~\cite{II2019Scalable}. As we all known, the emergence of large-scale  datasets with a diverse and general set of objects, like ImageNet DET~\cite{Deng-ImageNet-CVPR2009} and COCO~\cite{Lin2014COCO} has contributed greatly to rapid advances of object detection. As a special case of object detection, compared with  ImageNet DET~\cite{Deng-ImageNet-CVPR2009} and COCO~\cite{Lin2014COCO}, existing logo detection benchmarks lack a large number of categories and well-defined annotations.
\begin{figure}[!t]
	\centering
	\includegraphics[width=0.5\textwidth]{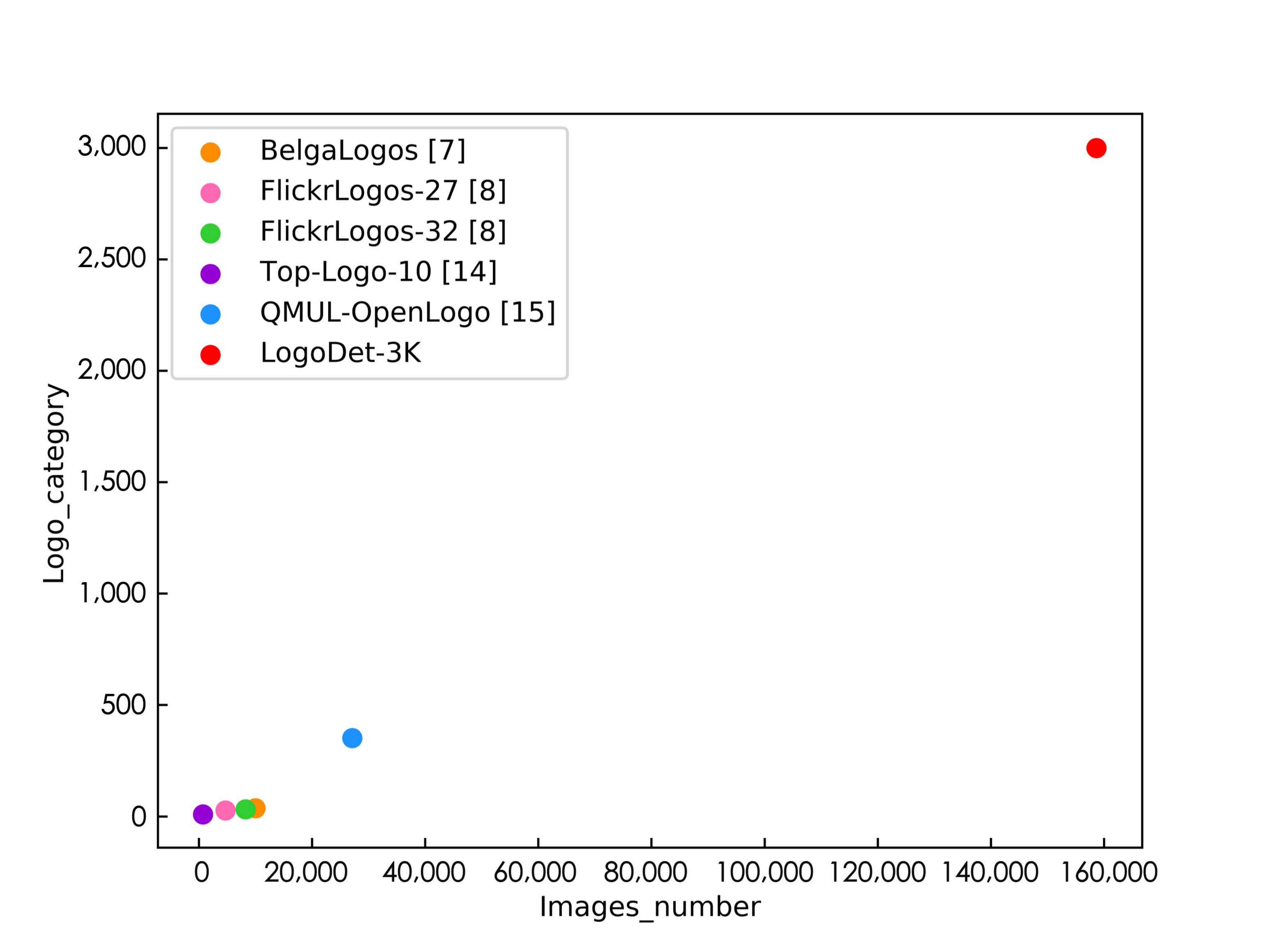}
	\caption{Statistics of LogoDet-3K categories and images. The abscissa represents the number of logo images, the ordinate represents the number of categories.}
	\label{comp_points_classes}
\end{figure}

\begin{figure*}[!t]
	\centering
	\includegraphics[width=1\textwidth]{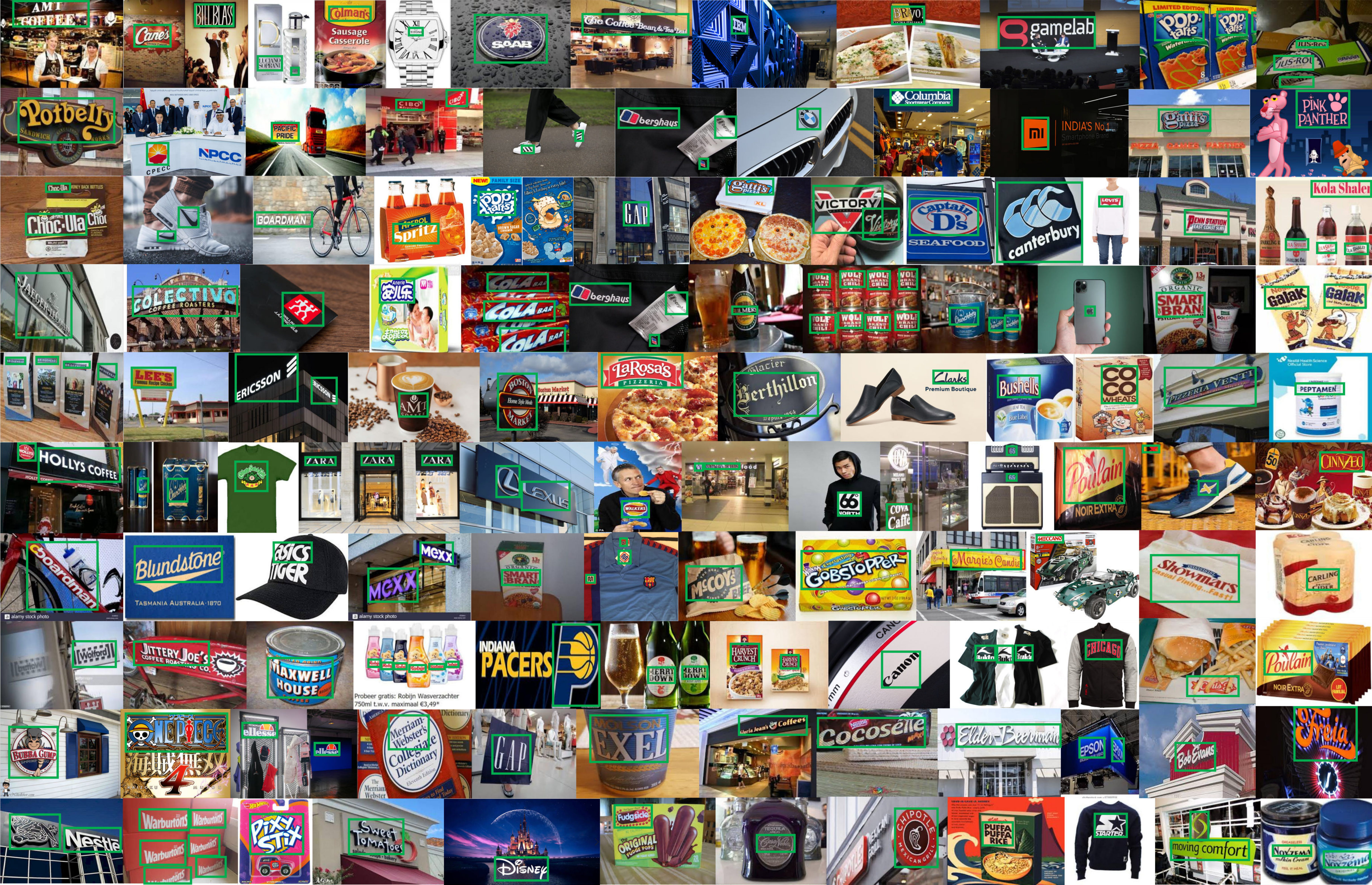}
	\caption{Image samples from various categories of LogoDet-3K.}
	\label{logo_category_samples}
\end{figure*}

Therefore, we introduce LogoDet-3K, a new large-scale, high-quality logo detection dataset. Compared with existing logo datasets, LogoDet-3K has three distinctive characteristics: (1) Large-scale. LogoDet-3K consists of 3,000 logo categories, 158,652 images and 194,261 bounding boxes. It has larger coverage on logo categories and larger quantity on annotated objects compared with existing logo datasets. (2) High-quality. Each image in the construction progress is strictly conformed to the pipeline which is carefully designed, including logo image collection, logo image filtering and logo object annotation. (3) High-challenge. Logo objects typically consist of mixed text and graphic symbols. Even the same logo can appear in different  scenarios such as various non-rigid, coloring and lighting transformations. For example, a rigid logo object when appearing in a real clothing image often becomes non-rigid, making it difficult to be detected. As shown in Fig.~\ref{comp_points_classes}, our proposed LogoDet-3K dataset far exceeds the existing logo dataset both in the number of categories and the number of images. Fig.~\ref{logo_category_samples} gives some image samples from various categories of LogoDet-3K. In addition, imbalanced samples and very small logo objects make this dataset more challenging.

We further propose a strong baseline method Logo-Yolo based on the network architecture YOLOv3 for logo detection. Logo-Yolo takes characteristics of LogoDet-3K, such as various logo object sizes, sample imbalance and different background scenarios into consideration, and incorporates Focal Loss~\cite{Tsung2017Focal} into the state-of-the-art detection framework YOLOv3 for logo detection. CIoU loss ~\cite{Zheng2020Distance} is further adopted to obtain more accurate regression results. Finally, we conduct comprehensive experiments on LogoDet-3K using several state-of-the-art object detection models and our proposed method, as well as ablation study and qualitative analysis.

This paper has three main contributions. (1) We introduce a new large-scale logo dataset  LogoDet-3K$\footnote{We will release the dataset upon publication.}$ with 3,000 classes, 194,261 objects and 158,652 images, which is the largest logo classes with full annotation.
(2) We propose a strong baseline method Logo-Yolo, which adopts the YOLOv3 detection framework, and combines Focal loss and CIoU loss to achieve better detection performance on LogoDet-3K. (3) We perform extensive experiments on LogoDet-3K by using several baseline models and our method, and further verify the effectiveness of our method and better generalization ability of LogoDet-3K on logo detection and retrieval tasks.

The rest of this paper is organized as follows. Section II reviews related work. Section III given the process of datasets construction and statistics. And Section IV elaborates the proposed large-scale logo detection method. Experimental results and analysis are reported in Section V. Finally, we conclude the paper and give future work in Section VI.

\section{Related Work}
Our work is closely related to two research fields:  (1) logo detection datasets and (2) logo detection researches.
\subsection{Logo Detection Datasets}
The large-scale dataset is an important factor for supporting advanced object detection algorithms, especially in the deep learning era, and it is no exception in logo detection. The first benchmark for logo detection is the BelgaLogos dataset~\cite{Neumann2001Integration}, which contains only 37 logo categories totaling 1,000 images. Over the years, some larger logo datasets such as FlickrLogos-32~\cite{Romberg2011Scalable} and Logos in the wild~\cite{Andras2017Open} have been proposed. However, these datasets lack the diversity and coverage in logo categories and images. For example, FlickrLogos-32 only consists of 32 logo categories with 70 images each category. This is far less than millions of images required in deep learning. Some researchers constructed some larger datasets, such as WebLogo-2M~\cite{Su2017WebLogo}, LOGO-Net~\cite{Hoi2015LOGO} and PL2K~\cite{II2019Scalable}. However, WebLogo-2M is collected from online search engines and just automatically be labeled at image level with  much noise, while PL2K and LOGO-Net are not publicly available.
\begin{table*}[!t]
	\caption{Comparison between LogoDet-3K and existing logo datasets.}
	\label{Logo_dataset_compare}
	\centering
	\setlength{\tabcolsep}{4mm}{
		\begin{tabular}{ccccccc}
			\hline
			$\sharp$Datasets        & $\sharp$Logos &$\sharp$Brands & $\sharp$Images    & $\sharp$Objects       & $\sharp$Supervision   & $\sharp$Public \\ \hline
			BelgaLogos~\cite{Neumann2001Integration}   & 37  &37      & 10,000    & 2,695   & Object-Level  & Yes          \\
			FlickrLogos-27~\cite{Romberg2011Scalable}  & 27  & 27      & 1,080     & 4,671   & Object-Level  & Yes          \\
			FlickrLogos-32~\cite{Romberg2011Scalable}  & 32  & 32      & 8,240     & 5,644   & Object-Level  & Yes          \\
			FlickrLogos-47~\cite{Romberg2011Scalable}  & 47  & 47       & 8,240     &  -      & Object-Level  & No           \\
			Logo-18~\cite{Hoi2015LOGO}                 & 18   &10     & 8,460     & 16,043  & Object-Level  & No           \\
			Logo-160~\cite{Hoi2015LOGO}                &160  & 100   & 73,414    & 130,608 & Object-Level  & No           \\
			Logos-32plus~\cite{Bianco2017Deep}         & 32   & 32   & 7,830    & 12,302 & Object-Level  & No           \\
			Top-Logo-10~\cite{Su2018Deep}              & 10   & 10  & 700       &  -      & Object-Level  & No           \\
			SportsLogo~\cite{Liao2017Mutual}           & 20   & 20  & 2,000     &  -      & Object-Level  & No           \\
			CarLogo-51~\cite{Car-51}&51&51&11903&-&Image-Level&No\\
			WebLogo-2M~\cite{Su2017WebLogo}            & 194   & 194    & 1,867,177 &  -      & Image-Level   & Yes          \\
			Logos-in-the-Wild~\cite{Andras2017Open}    & 871  & 871     & 11,054    & 32,850  & Object-Level  & Yes          \\
			QMUL-OpenLogo~\cite{Su2018Open}            & 352  & 352  & 27,083    &  -    & Object-Level  & Yes          \\
			PL2K~\cite{II2019Scalable}                 & 2,000  & 2,000   & 295,814   &  -    & Object-Level  & No           \\
			Logo-2K+~\cite{Wang2020Logo2K}             & 2,341  & 2,341   & 167,140   &  -          & Image-Level  & Yes           \\ \hline
			LogoDet-3K  &3,000  &2864   &158,652 &194,261 & Object-Level  &Yes        \\ \hline
	\end{tabular}}
\end{table*}

In order to solve the problem, we propose the LogoDet-3K, which is a large-scale, high-coverage and high-quantity dataset with 3,000 logo categories, 158,652 images and 194,261 objects.  Table \ref{Logo_dataset_compare} summarizes the statistics of existing logo datasets and LogoDet-3K. We can see that  LogoDet-3K has more logo categories and logo objects, which is more helpful to explore data-driven deep learning techniques for logo detection.

\subsection{Logo Detection}
In previous years, DPM~\cite{Pedro2010object} and HOG~\cite{Hoi2015LOGO}, are widely used as traditional object detection methods. Later, with the development of convolutional neural networks, more and more works start to utilize deep learning techniques, such as Faster RCNN~\cite{Ren2015},  YOLO~\cite{Joseph2018Yolov3} and~\cite{A2020Gao} self-attention for logo detection. In general, deep learning based object detector could be divided into two types: two-stage detector and single-stage detector. The popular two-stage detectors are the series of R-CNN like Faster RCNN~\cite{Ren2015}, which introduced the region proposal network and individual blocks to improve the detection performance. In contrast, the paradigm of single-stage detector aims to be faster and more efficient solution by classifying anchors directly and then refining them without proposal generation network, such as SSD~\cite{liu2016ssd}, RetinaNet~\cite{Tsung2017Focal} and YOLO series~\cite{Joseph2018Yolov3}. Recently, the proposed anchor-free method CornerNet~\cite{LawCornerNet2018} is highly acclaimed, while SNIPER~\cite{Singh2018SNIPER} and Cascade R-CNN~\cite{Cai2018CascadeR-CNN} are introduced to further improve the performance.

In general, logo detection has little advanced as a kind of generic object detection. An important reason is that the development of logo detection technology is limited by the size of logo dataset. Early logo detection methods are established on hand-crafted visual features (e.g. SIFT and HOG~\cite{Hoi2015LOGO}) and conventional classification models (e.g. SVM~\cite{Revaud2012Correlation}). Recently, some deep learning techniques have been applied in logo detection~\cite{Bianco2015Logo, Iandola2015DeepLogo, Kalantidis2011STL, Hang2020Scalable}. For example, Oliveira \emph{et al.}~\cite{Oliveira2016Automatic} adopted pre-trained CNN models and used them as a part of Fast Region-Based Convolutional Networks recognition pipeline. Feh{\'{e}}rv{\'{a}}ri \emph{et al.}~\cite{II2019Scalable} combined metric learning and basic object detection networks to achieve few-shot logo detection. Compared with existing logo detectors, our proposed Logo-Yolo is more effective for large-scale logo category and sample imbalance.

\section{LogoDet-3K}
\subsection{Dataset Construction}
The construction of LogoDet-3K is comprised of three steps, namely logo image collection, logo image filtering and logo object annotation. Each image is manually examined and reviewed to guarantee the quality of LogoDet-3K after filtering and annotation. The dataset building process is detailed in the following subsections. Additionally, each logo name is assigned to one of nine super-classes based on the daily need of life and the main positioning of common enterprises, namely Clothing, Food, Transportation, Electronics, Necessities, Leisure, Medicine, Sport and Others. In this paper, Table~\ref{super_class} gives the statistics of super classes of LogoDet-3K dataset.
\begin{table}[!t]
	\centering
	\caption{Data statistics on LogoDet-3K.}
	\label{super_class}
	\renewcommand\arraystretch{1.05}
	\setlength{\tabcolsep}{1mm}{
		\begin{tabular}{cccc}
			\hline
			Root-Category    & Sub-Category & Images & Objects\\ \hline
			Food	     &932	&53,350	  &64,276	\\
			Clothes      &604	&31,266	  &37,601	\\
			Necessities	 &432	&24,822	  &30,643	\\
			Others	     &371	&15,513	  &20,016	\\
			Electronic	 &224	&9,675	  &12,139	\\
			Transportation	&213	   &10,445	&12,791	\\
			Leisure	     &111	&5,685	  &6,573	\\
			Sports	     &66	&3,953	  &5,041	\\
			Medical	     &47	&3,945	  &5,185	\\
			\hline
			Total & 3,000&  158,652 &  194,261 \\ \hline
	\end{tabular}}
\end{table}

\textbf{Logo Image Collection.} A large-scale logo detection dataset should include comprehensive categories. Before crawling logo images, we built a comprehensive logo list based on the `Forbes Global 2,000'\footnote{\url{https://www.forbes.com/global2000/list/tab:overall}} and other famous logo  lists. Finally, we collected 3,000 logo names  for our logo vocabulary, which covers nine super-classes.

Subsequently, we used the logo name from the logo vocabulary as the query to crawl logo images from the Google search engine. Top-500 retrieved results were kept for the logo relevance for each query. In order to increase diversity of the dataset, we also crawled logo images from other online search engines including Bing and Baidu. In order to crawl more relevant images, we changed the search terms by adding `brand' or `logo' in search keywords. For example, there were so many images of shoes without any logo in the `Clarks' category, which is a famous British shoe company. We extended the search term such as `Clarks brand' or `Clarks logo' and obtained more relevant logo images as we expected.
\begin{figure*}[!t]
	\centering
	\includegraphics[width=0.85\textwidth]{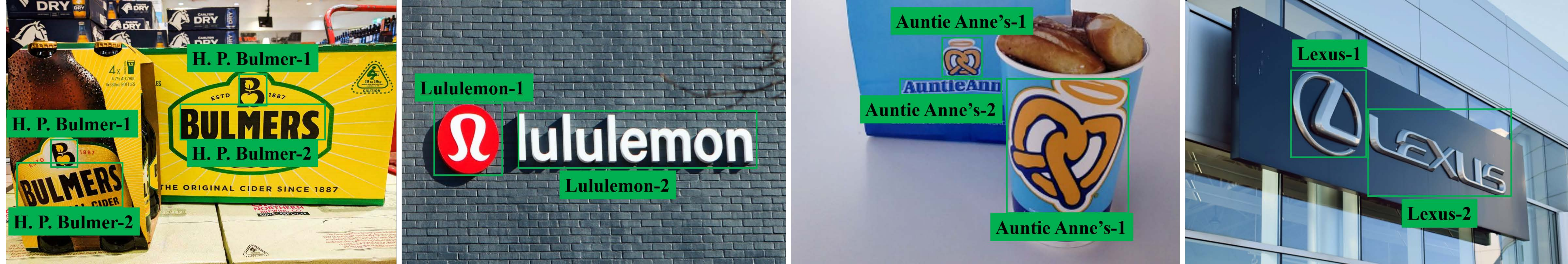}
	\caption{Multiple logo categories for some brands, where a distinction between these logo categories via adding the suffix `-1', `-2'.}
	\label{divided_category}
\end{figure*}
\begin{figure*}[!t]
	\centering
	\includegraphics[width=0.85\textwidth]{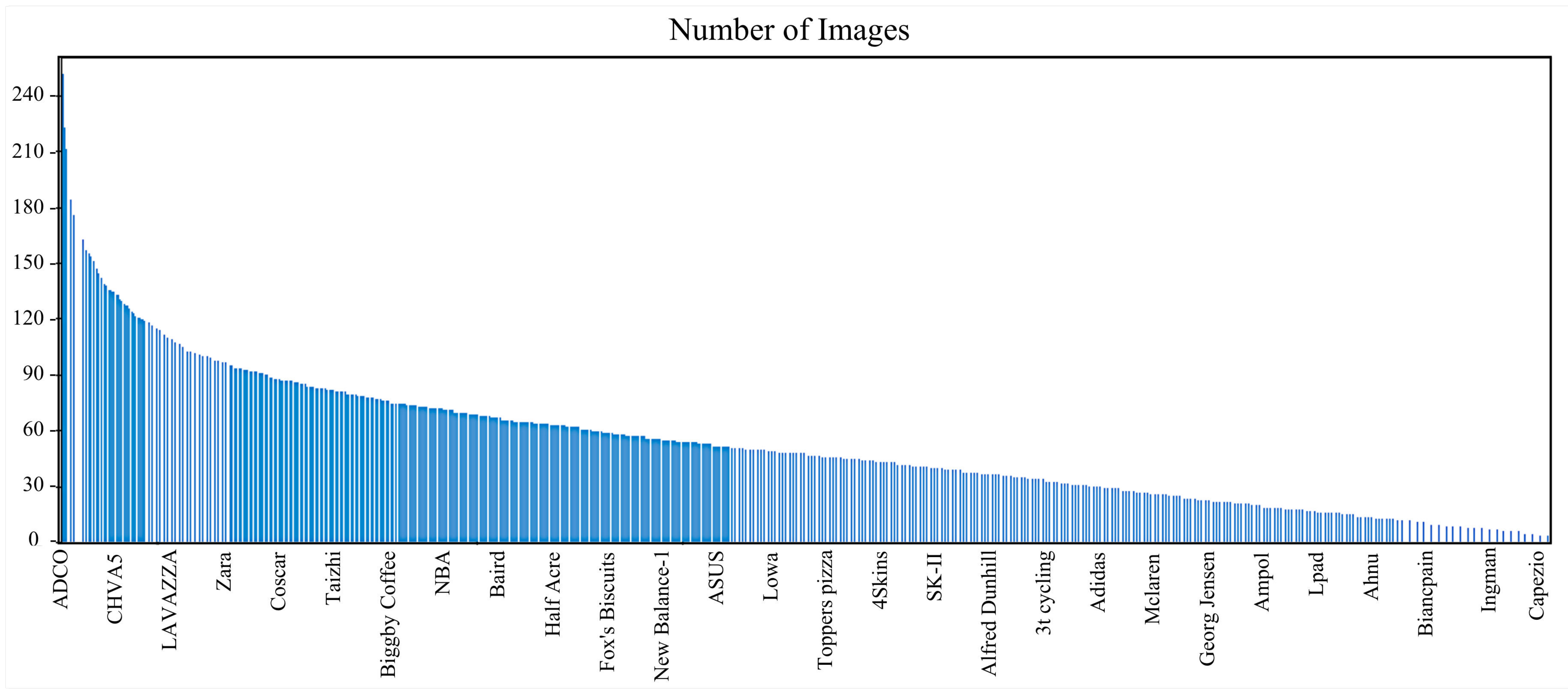}
	\caption{Sorted distribution of images for each logo in LogoDet-3K.}
	\label{statistic_histogram}
\end{figure*}

\textbf{Logo Image Filtering.} To guarantee the data quality, we cleaned the collected images manually before annotating them. Considering that not all the logo images are acceptable,  we check each logo category to guarantee that  it contained corresponding logo images with a suitable size and aspect ratio via  both automatic processing and manual cleaning. Particularly, we removed the following logo images, including: (1) images with  length or height  less than 300 pixels or extreme aspect ratio, (2) images with extreme aspect ratio, (3) duplicated images, (4) images without logos and (5) images with logos were not included in the logo vocabulary. In addition, a brand may have different types of logos, such as a symbolic logo and a textual logo or even more.  In this case, different types of logos should be treated as different logo categories for this brand similar to~\cite{Andras2017Open}. Fig.~\ref{divided_category} shows some examples, the suffix `-1', `-2' is added to the logo name as the new logo category, such as the `Lexus-1' presents the `Lexus' symbolic logo while `Lexus-2' presents its textual logo for the brand  `Lexus'.

\textbf{Logo Object Annotation.} As the most important step in constructing logo detection datasets, the annotation process takes a lot of time. The final annotation results follow some criterions. For example, if the logo is occluded, the annotators are instructed to draw the box around its visible parts. If an image contains multiple logo instances, each logo object needs to be annotated. In order to ensure the annotation quality of LogoDet-3K, each bounding box was annotated manually as close as possible to the logo object to avoid extra backgrounds. After finishing the above works, we inspected and examined all the annotated images labeled by the annotators. If an annotated image does not meet these requirements, the image will be rejected and need to be re-annotating.

\subsection{Dataset Statistics}
Our resulting LogoDet-3K consists of 3,000 logo classes, 158,652 images and 194,261 logo objects. To delve into the details of our dataset, we provide the statistics at the super-class and category level. Fig.~\ref{statistic_histogram} shows the distribution of images for each logo in LogoDet-3K. The thicker the columnar area in histogram, the larger the proportion. From Fig.~\ref{statistic_histogram}, we can see that imbalanced distribution across different logo categories are one characteristic of LogoDet-3K, posing a challenge for effective logo detection with few samples.

\begin{figure*}[!t]
	\centering
	\includegraphics[width=0.9\textwidth]{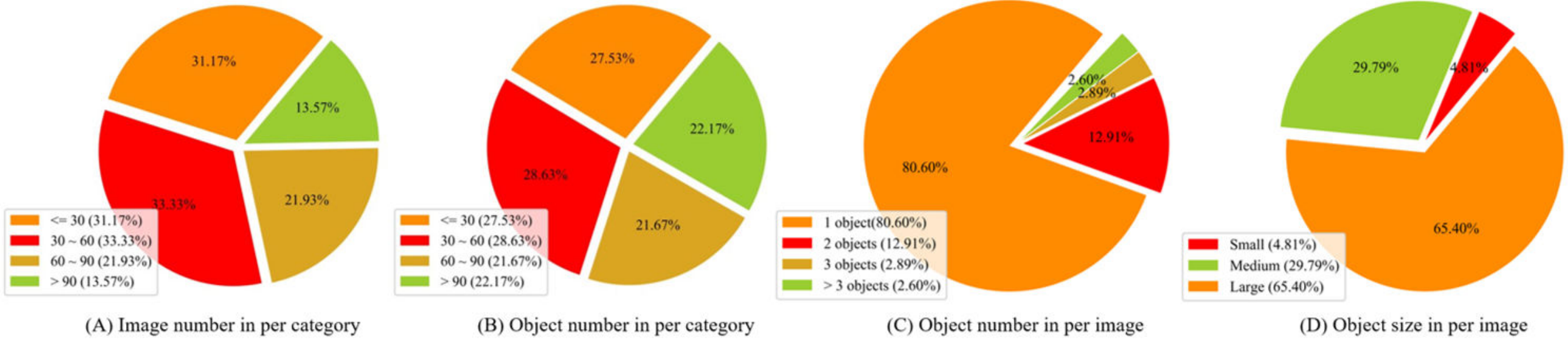}
	\caption{The detailed statistics of LogoDet-3K about Image and object distribution in per category, the number of objects in per image and object size in per image.}
	\label{statistic}
\end{figure*}

In addition, Fig.~\ref{statistic} summarizes the distribution of images and categories in LogoDet-3K. Fig.~\ref{statistic}~(A) shows the distribution of the number of images for each category. Fig.~\ref{statistic}~(B) shows the distribution of the number of objects of each class. As we can see,  there exists imbalanced distribution across different logo objects and images for different logo categories. Fig.~\ref{statistic}~(C) gives the number of objects in each image. We can see that most images contain  one or two logo objects. As shown in Fig.~\ref{statistic}~(D), LogoDet-3K is composed of  4.81\% small instances (area~\textless~32$^2$), 29.79\% medium instances (32$^2$~\textless=~area~\textless=~96$^2$) and 65.40\% large instances (area~\textgreater~96$^2$). The large percentage of  small and medium logo objects ($\sim$~35\%) will create another challenge to logo detection on this dataset, since small logos are harder to detect.

\begin{figure*}[!t]
	\centering
	\includegraphics[width=0.95\textwidth]{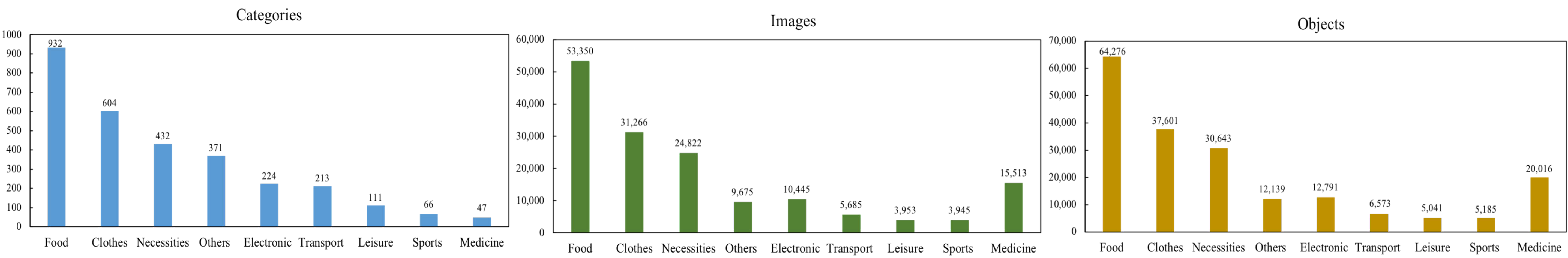}
	\caption{Distributions of categories, images and objects from LogoDet-3K on super-classes.}
	\label{super_class_statistics}
\end{figure*}

We also provide the statistics of logo categories, images and logo objects in 9 different super classes in Fig.~\ref{super_class_statistics}, which can direct to getting the difference on numbers. The Food, Clothes and Necessities class are larger in objects and images compared with other classes.



\section{Approach}
Taking characteristics of LogoDet-3K into consideration, we propose a strong baseline Logo-Yolo for logo detection, which  adopted the state-of-the-art deep detector YOLOv3 as the backbone to cope with small-scale  and multi-scale logos. Since the logo image contains fewer objects, there will be conducted more negative samples and hard samples, we utilized Focal Loss~\cite{Tsung2017Focal} to solve the problem of logo sample imbalance. In addition, we adopted K-means clustering statistics to re-compute the pre-anchors size for LogoDet-3K to select the best anchor size, and introduced recent proposed CIoU loss~\cite{Zheng2020Distance} to obtain more accurate regression results.

\textbf{Improved Losses for Logo Detection.} Fewer logo objects in the image produce more negative samples, leading to an imbalance between positive and negative samples. Focal Loss~\cite{Tsung2017Focal} is proposed  to solve the problem of sample imbalance. Therefore, we incorporates the Focal Loss into the whole loss of Logo-Yolo, the classification loss is formulated as follows:
\begin{equation}
\mathrm{Focal\;Loss}=\;\left\{\begin{array}{l}-\alpha{(1-y')}^\beta\log y'\;\;\;\;\;\;\;\;\;,\;y=1\\-(1-\alpha)y'^\beta\log(1-y')\;,\;y=0\end{array}\right.
\end{equation}
where $y\in\{\pm1\}$ is a ground-truth class and $y'\in\lbrack0,\;1\rbrack$ is the model's estimated probability by activation function. Focus loss introduces two factors $\alpha$ and $\beta$, where $\alpha$ is used to balance positive and negative samples, while $\beta$  focuses more on difficult samples.

In addition, $L_{n}$-norm loss is widely adopted for bounding box regression, while it is not tailored to the evaluation metric (Intersection over Union (IoU)) in existing methods. We further incoporate the CIoU loss~\cite{Zheng2020Distance} into the whole loss of YOLOv3 to solve the problem of inconsistency between the metric and the border regression on logo detection, and the IoU-based loss can be defined as,
\begin{equation}
L_{CIoU}=1-IoU+R_{CIoU}(B_{pd},B_{gt})
\end{equation}
where $R_{CIoU}$ is penalty term for predicted box $B_{pd}$ and target box $B_{gt}$.  

CIoU loss considered three geometric factors in the bounding box regression, including overlap area, central point distance and aspect ratio to solve the problem of inconsistency between the metric and the border regression during logo detection. Therefore, the method to minimize the normalized distance between central points of two bounding boxes, and the penalty term can be defined as,
\begin{equation}
R_{CIoU}=\frac{\varphi^2(b,b_{gt})}{c^2}+\alpha\frac4{\mathrm\pi^2}{(arc\tan\frac{w^{gt}}{h^{gt}}-arc\tan\frac wh)}^2	
\end{equation}
where $b$ and $b_{\mathbf g\mathbf t}$ denote the central points of $B_{pd}$ and $B_{gt}$, $\varphi(\cdot)$ is the Euclidean distance, and $c$ is the diagonal length of the smallest enclosing box covering the two boxes. $\alpha$ is a positive trade-off parameter. $w$, $h$ are aspect ratio of the prediected box, respectively.

\textbf{Pre-anchors Design for Logo Detection.} Anchor boxes are a set of initial fixed width-and-height candidate boxes. Those defined by the original network are no longer suitable for LogoDet-3K. Therefore, we use K-means clustering algorithm to perform clustering analysis on the bounding boxes for objects of LogoDet-3K and then select the average overlap degree (Avg IoU) as the metric for clustering result analysis.  We can obtain the number of anchor boxes based on the relationship between the number of samples and Avg IoU.

The aggregated Avg IoU objective function $f$ can be expressed as,
\begin{equation}
f=\mathrm{argmax}\frac{{\displaystyle\sum_{i=1}^k}{\displaystyle\sum_{j=1}^{N_k}}I_{IoU}\left(B,\;C\right)}N
\end{equation}
where $B$ represents the ground-truth sample and $C$ represents the center of the cluster. $N$ represents the total number of samples, $k$ represents the number of clusters. In general, we adopt the K-means clustering algorithm to select the number of candidate anchor boxes and aspect ratio dimensions.

\section{Experiment}
\subsection{Experimental Setup}
For parameter settings, we design pre-anchor boxes for different object detectors via calculations on LogoDet-3K dataset. In our method, the number of anchor boxes is set as 9, according to the relationship between the number of samples and Avg IoU via K-means clustering. The final results of anchor centers are (53, 35), (257, 151), (75, 104), (271, 248), (159, 118), (134, 220), (270,73), (115, 46) and (193, 58), which are width and height of the corresponding cluster centers on the LogoDet-3K dataset. For the Focal loss of Logo-Yolo, $\alpha=0.25$, $\beta=2$. 

For the evaluation metric, we use mean Average Precision (mAP)~\cite{Everingham2010voc} and the IoU threshold is 0.5, which means that a detection will be considered as positive if the IoU between the predicted box and ground-truth box exceeds 50\%. 
\begin{table} [!t]
	\caption{Statistics of three benchmarks.}
	\label{dataset comparison}
	\centering
	\setlength{\tabcolsep}{0.3mm}{
		\begin{tabular}{cccccccc}
			\hline
			$\sharp$Datasets&$\sharp$Classes&$\sharp$Images&$\sharp$Objects&$\sharp$Trainval&$\sharp$Test\\
			\hline
			LogoDet-3K-1000 &1,000&85,344&101,345&75,785&11,236\\
			LogoDet-3K-2000 &2,000&116,393&136,815&103,356&13,037\\
			LogoDet-3K &3,000&158,652&194,261&142,142&16,510\\
			\hline
	\end{tabular}}
\end{table}
\begin{table} [!t]
	\caption{Statistics of three super-classes.}
	\label{super-classes comparison}
	\centering
	\setlength{\tabcolsep}{1.5mm}{
		\begin{tabular}{cccccccc}
			\hline
			$\sharp$Datasets&$\sharp$Classes&$\sharp$Images&$\sharp$Objects&$\sharp$Trainval&$\sharp$Test\\
			\hline	
			Food	     &932	&53,350	  &64,276	&47,321 &6,029\\
			Clothes      &604	&31,266	  &37,601	&27,732 &3,534\\
			Necessities	 &432	&24,822	  &30,643	&22,017 &2,805\\
			\hline
	\end{tabular}}
\end{table}

For the experiment datasets, we define various data subsets as different benchmarks by means of random division on the overall LogoDet-3K dataset. Particularly, we divide the LogoDet-3K dataset into three subsets including 1,000, 2,000 and 3,000 categories, respectively. Through those experiments, we verify the robustness of our method as the number of categories and images increases. The statistics of three sub-datasets are shown in Table~\ref{dataset comparison}. In addition, we conduct experiments based on super categories. The categories with the largest number of the three categories are also common logo categories in real world, including Food, Clothes, and Necessities. This experiment is to explore the detection effect of our method on common categories and the characteristics of the three categories of datasets. The statistics of three subsets from these super categories are shown in Table \ref{super-classes comparison}. 

Experiments are performed with state-of-the-art object detectors: Faster R-CNN~\cite{Ren2015}, SSD~\cite{liu2016ssd}, RetinaNet~\cite{Tsung2017Focal}, FPN~\cite{Lin2017FPN}, Cascade R-CNN~\cite{Cai2018CascadeR-CNN}, Distance-IoU~\cite{Zheng2020Distance} and YOLOv3~\cite{Joseph2018Yolov3}. For their backbones, we adopt the general setting: ResNet101 is selected as the backbone for Faster R-CNN, RetinaNet, FPN and Cascade R-CNN. Darknet-53 is used as the backbone of YOLOv3 and Distance-IoU, and VGGNet-16~\cite{VGG2015} for SSD. The experiments are conducted in PyTorch and DarkNet framework, GPU with the NVIDIA Tesla K80 and Tesla V100.
\begin{figure*}[!t]
	\centering
	\includegraphics[width=0.93\textwidth]{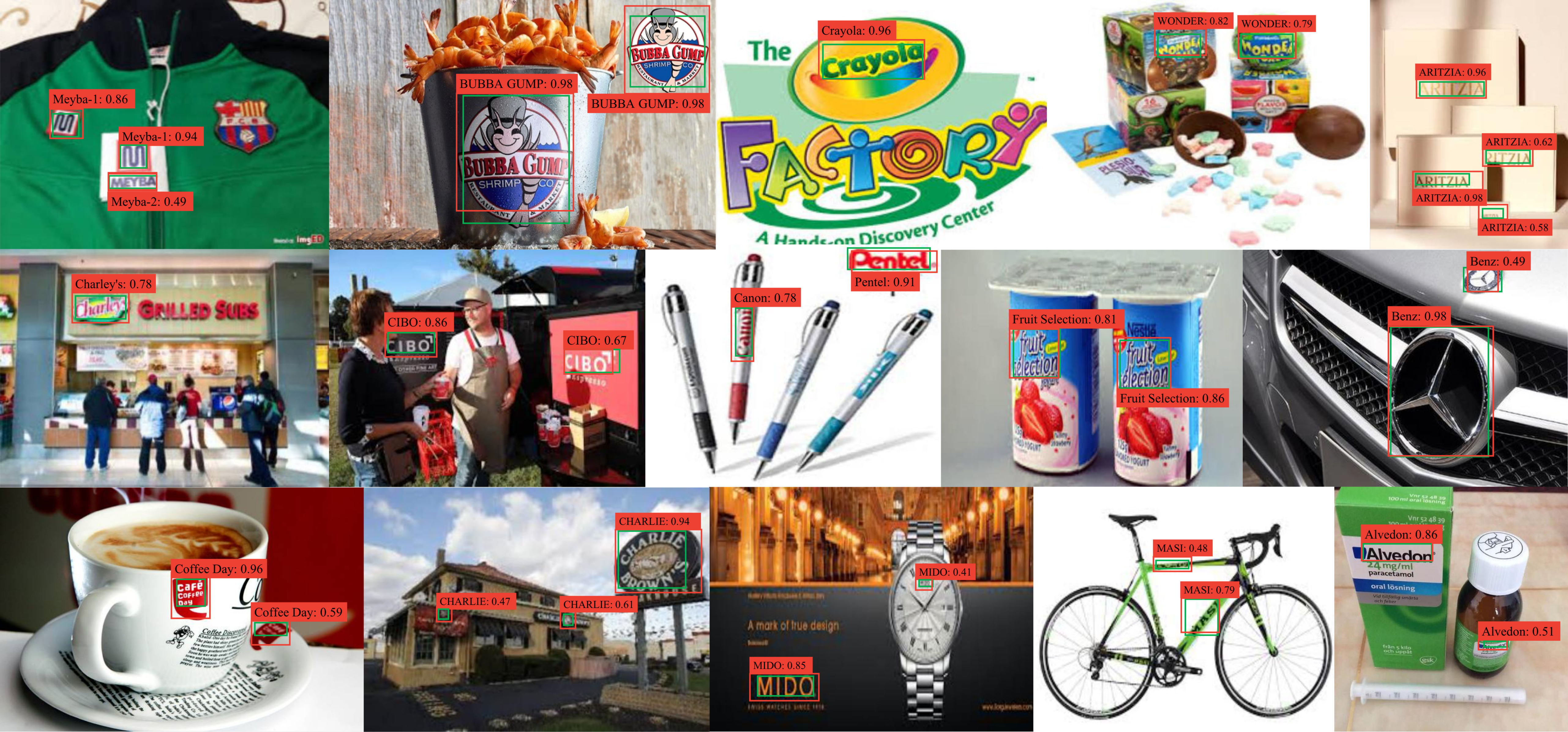}
	\caption{Some detection results of Logo-Yolo on LogoDet-3K.}
	\label{result}
\end{figure*}

\begin{table}[!t]
	\caption{Comparison of baselines on different benchmarks (\%).}
	\label{Result}
	\centering
	\footnotesize
	\setlength{\tabcolsep}{3mm}{
		\begin{tabular}{ccccc}
			\hline Benchmarks                & Methods                              &Backbones           & mAP \\ \hline
			\multirow{8}{*}{LogoDet-3K-1000} & Faster RCNN~\cite{Ren2015}           & ResNet-101         & 45.16            \\
			& SSD~\cite{liu2016ssd}                & VGGNet-16          & 43.32            \\
			& RetinaNet~\cite{Tsung2017Focal}      & ResNet-101         & 52.10            \\
			& FPN~\cite{Lin2017FPN}                                  & ResNet-101         & 49.63            \\
			& Cascade R-CNN\cite{Cai2018CascadeR-CNN}                        & ResNet-101         & 48.14            \\  
			& Distance-IoU~\cite{Zheng2020Distance}                         & DarkNet-53         & 53.06    \\                                                      
			& YOLOv3~\cite{Joseph2018Yolov3}       & DarkNet-53         & 55.21            \\
			& \textbf{Logo-Yolo}                   & \textbf{DarkNet-53}         & \textbf{58.86}   \\ \hline
			\multirow{8}{*}{LogoDet-3K-2000} & Faster RCNN~\cite{Ren2015}           & ResNet-101         & 41.86            \\
			& SSD~\cite{liu2016ssd}                & VGGNet-16          & 38.97         \\
			& RetinaNet~\cite{Tsung2017Focal}      & ResNet-101         & 49.00            \\
			& FPN~\cite{Lin2017FPN}                                  & ResNet-101         & 47.91            \\
			& Cascade R-CNN\cite{Cai2018CascadeR-CNN}                        & ResNet-101         & 46.32             \\     
			& Distance-IoU~\cite{Zheng2020Distance}                         & DarkNet-53         & 51.69    \\                                        
			& YOLOv3~\cite{Joseph2018Yolov3}       & DarkNet-53         & 52.32            \\
			& \textbf{Logo-Yolo}                   & \textbf{DarkNet-53}         & \textbf{56.42}  \\ \hline
			\multirow{8}{*}{LogoDet-3K} & Faster RCNN~\cite{Ren2015}                & ResNet-101         & 38.30            \\
			& SSD~\cite{liu2016ssd}                & VGGNet-16          & 34.47          \\
			& RetinaNet~\cite{Tsung2017Focal}      & ResNet-101         & 44.32            \\
			& FPN~\cite{Lin2017FPN}                                  & ResNet-101         & 42.84            \\
			& Cascade R-CNN\cite{Cai2018CascadeR-CNN}                        & ResNet-101         & 41.23            \\ 
			& Distance-IoU~\cite{Zheng2020Distance}                         & DarkNet-53         & 46.34    \\                                            
			& YOLOv3~\cite{Joseph2018Yolov3}       & DarkNet-53         & 48.61            \\
			& \textbf{Logo-Yolo}                   & \textbf{DarkNet-53}         & \textbf{52.28}   \\ \hline
	\end{tabular}}
\end{table}

\begin{table}[!t]
	\caption{Comparison of super-classes on different methods (\%).}
	\label{super}
	\centering
	\footnotesize
	\setlength{\tabcolsep}{3mm}{
		\begin{tabular}{cccc}
			\hline Benchmarks                & Methods                              &Backbones           & mAP \\ \hline
			\multirow{8}{*}{Food} & Faster RCNN~\cite{Ren2015}                      & ResNet-101         &47.32        \\
			& SSD~\cite{liu2016ssd}                & VGGNet-16          &46.18             \\
			& RetinaNet~\cite{Tsung2017Focal}      & ResNet-101         &51.46             \\
			& FPN~\cite{Lin2017FPN}                                  & ResNet-101         &51.10             \\
			& Cascade R-CNN\cite{Cai2018CascadeR-CNN}                        & ResNet-101         &52.46             \\  
			& Distance-IoU~\cite{Zheng2020Distance}                         & DarkNet-53         &53.11     \\                                                      
			& YOLOv3~\cite{Joseph2018Yolov3}       & DarkNet-53         &53.49             \\
			& \textbf{Logo-Yolo}                   & \textbf{DarkNet-53}         & \textbf{56.73}   \\ \hline
			\multirow{8}{*}{Clothes} & Faster RCNN~\cite{Ren2015}                   & ResNet-101         &51.63             \\
			& SSD~\cite{liu2016ssd}                & VGGNet-16          &49.74         \\
			& RetinaNet~\cite{Tsung2017Focal}      & ResNet-101         &55.98             \\
			& FPN~\cite{Lin2017FPN}                                  & ResNet-101         &55.62             \\
			& Cascade R-CNN\cite{Cai2018CascadeR-CNN}                        & ResNet-101         &56.90              \\     
			& Distance-IoU~\cite{Zheng2020Distance}                         & DarkNet-53         &56.54     \\                                        
			& YOLOv3~\cite{Joseph2018Yolov3}       & DarkNet-53         &57.01             \\
			& \textbf{Logo-Yolo}                   & \textbf{DarkNet-53}         & \textbf{61.32}  \\ \hline
			\multirow{8}{*}{Necessities} & Faster RCNN~\cite{Ren2015}                & ResNet-101        &52.22        \\
			& SSD~\cite{liu2016ssd}                & VGGNet-16          &50.03         \\
			& RetinaNet~\cite{Tsung2017Focal}      & ResNet-101         &54.01             \\
			& FPN~\cite{Lin2017FPN}                                  & ResNet-101         &53.37             \\
			& Cascade R-CNN\cite{Cai2018CascadeR-CNN}                        & ResNet-101         &55.49            \\ 
			& Distance-IoU~\cite{Zheng2020Distance}                         & DarkNet-53         &57.20     \\                                            
			& YOLOv3~\cite{Joseph2018Yolov3}       & DarkNet-53         &57.68             \\
			& \textbf{Logo-Yolo}                   & \textbf{DarkNet-53}         & \textbf{61.43}   \\ \hline
	\end{tabular}}
\end{table}

\begin{figure*}[!t]
	\centering
	\includegraphics[width=0.95\textwidth]{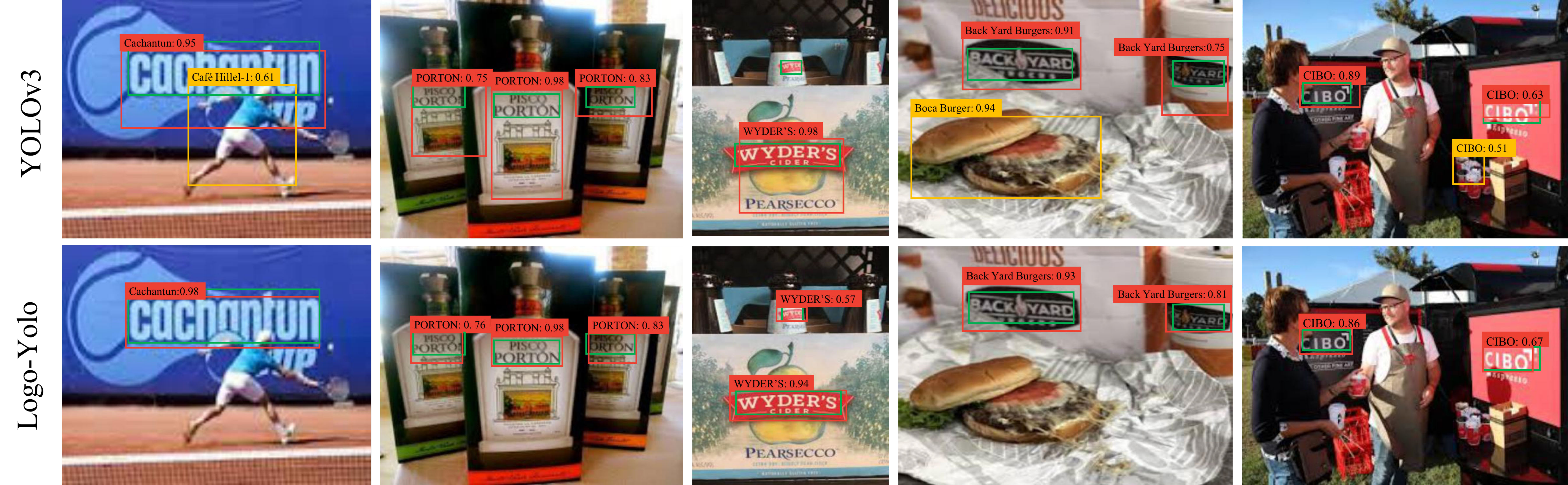}
	\caption{Qualitative result comparison on LogoDet-3K between YOLOv3 and Logo-Yolo. Green boxes:  ground-truth boxes. Red boxes: correct detection boxes. yellow boxes: mistakes detection boxes.}
	\label{compare}
\end{figure*}
\subsection{Experimental Results}
Table~\ref{Result} summarizes the results on three subsets among different detection models. Compared with existing baselines Faster RCNN, SSD and RetinaNet etc., YOLOv3 detector obtains better results on three subsets, which are 55.21\%, 52.32\% and 48.60\% respectively. The results of YOLOv3 are higher than Faster RCNN detector, because there are more small logo objects and fewer objects for many images in real-world scenarios, and the one-stage method is more suitable for this case. Therefore, we use the one-stage YOLOv3 detector as the basis of our method.

We then compare the performance of Logo-Yolo with all baselines, and observe that Logo-Yolo achieves the best performance among these models. It's worth noting that mAP of Logo-Yolo is 58.86\%, 56.42\% and 52.28\% on three benchmarks, and Logo-Yolo achieves the performance gain with 3.65\%, 4.10\% and 3.67\% compared with YOLOv3 in Table~\ref{Result}. Our method Logo-Yolo detection performance achieves the best result on the 1000-2000-3000 datasets, which proves the stability of the method.

Some detection results of Logo-Yolo are given in Fig.~\ref{result}, including the regression bounding box and the classification accuracy. The red box represents the prediction box and the green box is the ground-truth box. Clearly, Logo-Yolo can detect objects with occlusion, ambiguities and smaller, it obtains more accurate bounding box regression. And as shown in Fig.~\ref{compare}, the detector YOLOv3 makes some detection mistakes, such as treating a person or hamburger as logos, and thus the bounding boxes of detected logos are inaccurate, or missing. In contrast, our method obtains better performance both in the bounding box regression and the confidence of detected logos. In particular, our method has an advantage in small logo detection, such as the detected logos in the last two images in Fig.~\ref{compare}. 
\begin{figure*}[!t]
	\centering
	\includegraphics[width=0.9\textwidth]{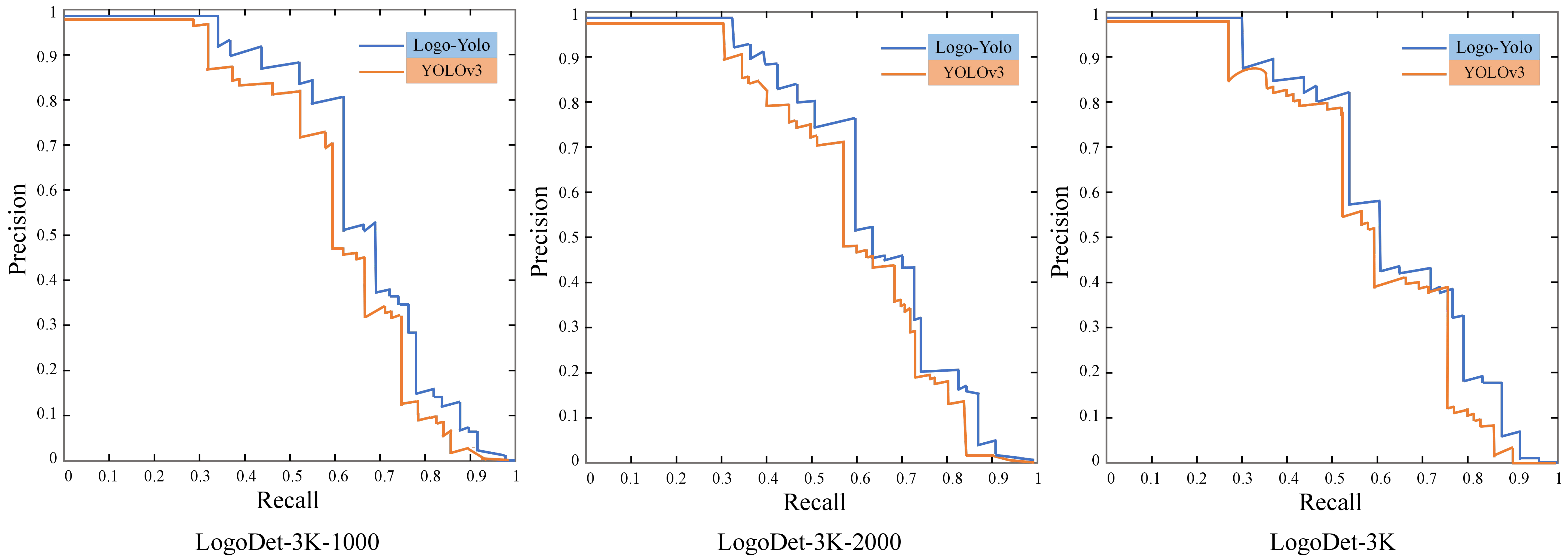}
	\caption{The Precision-Recall curve of Logo-Yolo and YOLOv3. The larger the enclosing area under the curve, the better the detection effect.}
	\label{PR}
\end{figure*}
\begin{figure*}[!t]
	\centering
	\includegraphics[width=0.9\textwidth]{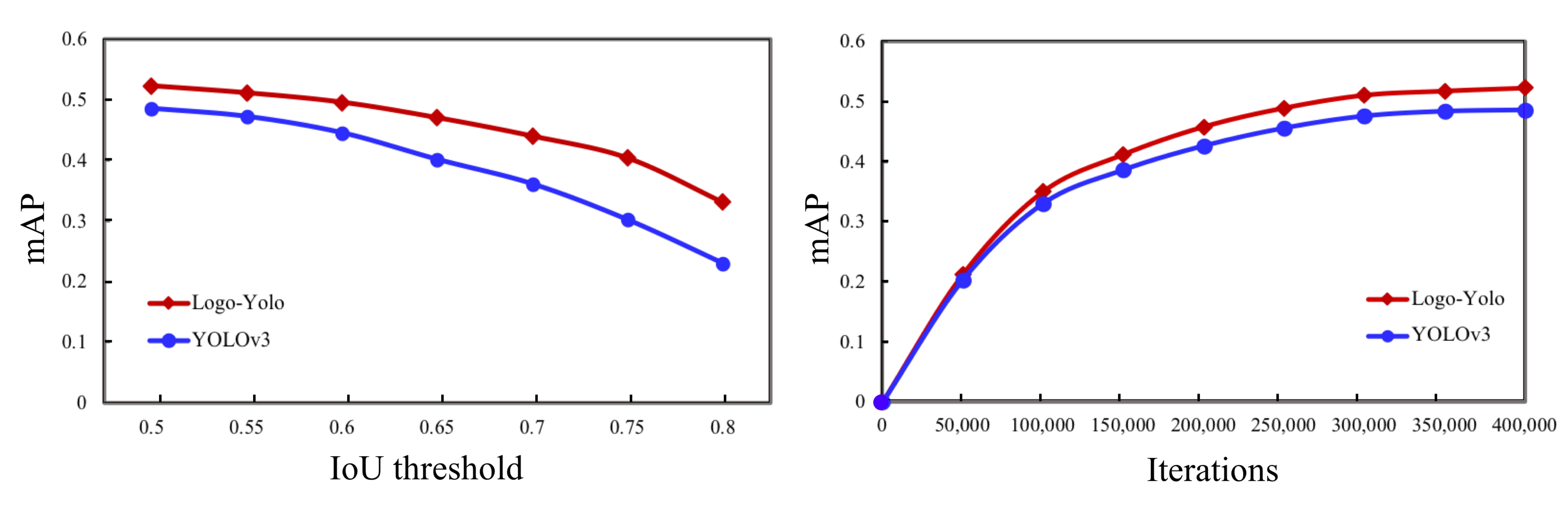}
	\caption{Left: Performance evaluation for different IoU thresholds. Right: The comparison of Logo-Yolo and YOLOv3 with increasing iterations.}
	\label{parameter_sensitivity}
\end{figure*}

In addition, Table~\ref{super} gives the comparison of three super-classes on different methods. Compared with existing baselines, the Logo-Yolo detector also obtains better results with 56.73\%, 61.32\% and 61.43\% on the super classes of Food, Clothes, and Necessities, respectively, which are 3.24\%, 4.31\% and 3.75\% higher than YOLOv3. This experiment also illustrates the effectiveness of our method. As we can see from Table~\ref{super}, the number of Necessities categories is 172 less than the clothes categories, but relatively similar detection results have been obtained (61.32\% vs 61.43\%), indicating that the Necessities category dataset is more difficult to detect. Analyzing food logos with a large number of categories and images, the detection performance of the 932 food category is slightly lower than the 1000 subset (56.73\% vs 58.86\%). The result shows that food-related logo detection is more challenging.

\subsection{Analysis}
Since Logo-Yolo and YOLOv3 obtain better detection performance, we next focus on the analysis via the comparison between two methods.

\textbf{Dataset Scale.} According to Table~\ref{Result}, the drop of Logo-Yolo in mAP is 2.44\% and 4.14\% when the number of categories increases from 1,000 to 2,000 and 2,000 to 3,000. Compared with YOLOv3, our model achieves better performance than other baselines on datasets with different scales, which proves a higher robustness on LogoDet-3K. We further calculate the Precision and Recall to illustrate the accuracy and missed detection rate. We use the Precision-Recall curve to show the trade-off between Precision and Recall in Fig.~\ref{PR} between YOLOv3 and Logo-Yolo. The larger the enclosing area under the curve, the better the detection performance. As shown in Fig.~\ref{PR}, Logo-Yolo has significantly improved the recall rate, which indicates that our method alleviated the problem of missing small objects  in logo detection.

\textbf{Parameter Sensitivity.} We evaluate the performance by varying different IoU thresholds from 0.5 to 0.8 at an interval of 0.05. As shown in Fig.~\ref{parameter_sensitivity} (Left), Logo-Yolo (red curve) has a more stable performance improvement than YOLOv3 (blue curve) when changing the IoU threshold. We also set different iterations to compare the convergence and accuracy of models. Fig.~\ref{parameter_sensitivity} (Right) shows higher performance with increasing iterations. It can be seen that our method converges at about 400,000 iterations and keeps higher accuracy than YOLOv3 in the training process.
\begin{table}[!t]
	\caption{Evaluation on individual modules and two modules of Logo-Yolo (\%).}
	\label{Ablation}
	\centering
	\setlength{\tabcolsep}{3mm}{
		\begin{tabular}{ccc}
			\hline
			Model                          & mAP \\ \hline
			YOLOv3                    &48.61   \\
			YOLOv3+Pre-anchors Design  &50.12   \\
			YOLOv3+Focal Loss          &49.21  \\
			YOLOv3+CIoU loss           &49.86   \\
			Logo-Yolo(w/o Pre-anchors Design)  &49.92  \\
			Logo-Yolo(w/o Focal Loss) &51.50   \\
			Logo-Yolo(w/o CIoU loss)  &50.64  \\
			\textbf{Logo-Yolo}                 &\textbf{52.28}  \\ \hline
	\end{tabular}}
\end{table}

\subsection{Ablation Study}
We conduct a comprehensive analysis of effects of three sub-variables and two modules from Logo-Yolo. Table~\ref{Ablation} shows an ablation study on the effects of different combinations of K-means, Focal Loss and CIoU loss. Firstly, three modules are added to YOLOv3, and the results improve 1.51\%, 0.60\% and 1.25\%, which proves the effectiveness of the Pre-anchors Design, Focal Loss and CIoU loss, respectively. Then, we conduct the two modules experiments from Logo-Yolo. The result of Logo-Yolo is higher than Logo-Yolo without Pre-anchors Design, which explains the effectiveness of two losses. Similarly, compared to Logo-Yolo without Focal Loss or CIoU loss, our proposed method achieves improvement, which demonstrates the effectiveness of another two modules for Logo-Yolo.

\subsection{Generalization Ability on Logo Detection}
To evaluate the robustness and generalization ability of Logo-Yolo architecture and its pre-trained models, we explore other two datasets Top-Logo-10~\cite{Su2018Deep} and FlickrLogos-32~\cite{Romberg2011Scalable}. The former contains 10 unique logo classes with 70 images for each logo class, and the latter is a popular logo dataset with full annotations, comprising 8,240 images from 32 categories. Logo-Yolo (per-trained) first loades the model trained on LogoDet-3K, and is then trained on the target dataset while Logo-Yolo is directly trained on the target dataset with random parameter initialization.
\begin{table}[!t]
	\caption{The performance of Logo-Yolo on Top-Logo-10 (\%).}
	\label{10}
	\centering
	\setlength{\tabcolsep}{8mm}{
		\begin{tabular}{cc}
			\hline
			Method                     & mAP   \\ \hline
			Faster RCNN~\cite{Ren2015} & 41.80  \\        
			SSD~\cite{liu2016ssd}      & 38.70  \\  
			YOLO~\cite{Redmon2016YOLO} & 44.58   \\  
			YOLOv3~\cite{Joseph2018Yolov3}  & 50.10      \\
			\textbf{Logo-Yolo}  & \textbf{52.17}      \\
			\textbf{Logo-Yolo (Pre-trained)} & \textbf{53.62} \\ \hline
	\end{tabular}}
\end{table}

\begin{table}[!t]
	\caption{The performance of Logo-Yolo on FlickrLogos-32 (\%).}
	\label{32}
	\centering
	\setlength{\tabcolsep}{8mm}{
		\begin{tabular}{cc}
			\hline
			Method                                       & mAP \\ \hline
			Bag of  Words (BoW)~\cite{Romberg2013Bundle} & 54.50   \\ 
			Deep Logo~\cite{Iandola2015DeepLogo}         & 74.40   \\  
			BD-FRCN-M~\cite{Oliveira2016Automatic}       & 73.50   \\   
			Faster RCNN~\cite{Ren2015}                   & 70.20   \\   
			YOLO~\cite{Redmon2016YOLO}                   & 68.70  \\
			YOLOv3~\cite{Joseph2018Yolov3}               & 71.70  \\
			\textbf{Logo-Yolo}                           & \textbf{74.62} \\
			\textbf{Logo-Yolo (Pre-trained)} & \textbf{76.11} \\ \hline
	\end{tabular}}
\end{table}
Table~\ref{10} summarizes experimental results for Top-Logo-10. We observe that our method Logo-Yolo achieves better performance compared with other models. There is further about 1.5 percent improvement after pre-training on LogoDet-3K, showing better generalization ability of LogoDet-3K. We can also see similar trends on FlickrLogo-32 in Table~\ref{32}. Overall, the evaluation on these two datasets verify the effectiveness of Logo-Yolo, and also shows better generalization ability of LogoDet-3K on other logo detection datasets.

In addition, we further select QMUL-OpenLogo dataset to evaluate the general object detection. This dataset is the largest publicly available logo detection dataset, and contains 352 categories and 27,038 images. To further exploit the fine-tuning capability of LogoDet-3K, we analyze the difference between LogoDet-3K pre-trained weights and QMUL-OpenLogo pre-trained weights. 

According to Table~\ref{QMUL}, our LogoDet-3K dataset shows strong generalization ability. Compared with YOLOv3 and Logo-Yolo method, our fine-tuned LogoDet-3K model for QMUL-OpenLogo detection can significantly boost the performance, with 1.73 points (53.69\% vs 51.96\%) for YOLOv3, and 2.16 points (55.37\% vs 53.21\%) for Logo-Yolo, the Logo-Yolo gains a 1.68 improvement (55.37\% vs 53.69\%). The results are shown that the effectiveness of pre-trained models and Logo-Yolo method. By pre-training the LogoDet-3K dataset which removes the 352 categories from QMUL-OpenLogo (LogoDet-3K w/o QMUL-OpenLogo), we can still achieve competitive results with 52.36\% on the QMUL-OpenLogo benchmark, 0.4 points higher than the result in YOLOv3 method, and 1.25 points for Logo-Yolo. It shows that the LogoDet-3K dataset has the generalization ability. Compared with QMUL-OpenLogo, our LogoDet-3K benchmark has much higher performance gain. By involving QMUL-OpenLogo Pre-training before LogoDet-3K, we can slightly improve the YOLOv3 with 0.34. For the Logo-Yolo, the QMUL-OpenLogo pre-training before LogoDet-3K can further bring in 0.73 points gain. The results shows LogoDet-3K contains richer logo features than QMUL-OpenLogo dataset, which can be widely used for logo detection.
\begin{table}[!t]
	\caption{Generalization ability of general object detection results on the QMUL-OpenLogo dataset (\%).}
	\label{QMUL}
	\small
	\centering
	\setlength{\tabcolsep}{0.5mm}{
		\begin{tabular}{ccc}
			\hline
			Method                    &Pre-trained Dataset & mAP \\ \hline
			YOLO9000~\cite{Redmon2017Yolo9000}      & QMUL-OpenLogo    & 26.33  \\ 
			YOLOv2+CAL~\cite{Su2018Open}            & QMUL-OpenLogo    & 49.17  \\
			FR-CNN+CAL~\cite{Su2018Open}            & QMUL-OpenLogo    & 51.03  \\
			YOLOv3                    & QMUL-OpenLogo    & 51.96     \\ 
			YOLOv3                    & LogoDet-3K w/o QMUL-OpenLogo    & 52.36  \\ 
			YOLOv3                    & LogoDet-3K    &  53.69 \\ 
			\textbf{YOLOv3}                    & \textbf{QMUL-OpenLogo -\textgreater~ LogoDet-3K}    & \textbf{54.03}  \\ \hline
			Logo-Yolo                 & QMUL-OpenLogo    & 53.21  \\ 
			Logo-Yolo                 & LogoDet-3K w/o QMUL-OpenLogo    & 54.46  \\ 
			Logo-Yolo                 & LogoDet-3K    & 55.37  \\ 
			\textbf{Logo-Yolo}                 & \textbf{QMUL-OpenLogo -\textgreater~ LogoDet-3K}    & \textbf{56.10}  \\ \hline
	\end{tabular}}
\end{table}

\subsection{Generalization Ability on Logo Retrieval}
For the retrieval experiments, each of the ten FlickrLogos-32 train samples for each brand serves as query sample. This allows to assess the statistical significance of results similar to a 10-fold-cross-validation strategy. As shown in Table~\ref{Retrieval} the ResNet101+Litw~\cite{Andras2017Open} is the better logo retrieval method. Detected logos are described by the feature extraction network outputs where three different state-of-the-art classification architectures, namely VGG16, ResNet101 and DenseNet161, serve as base networks in Table~\ref{Retrieval}. In addition, we adopt the proposed method Logo-Yolo to FlickrLogos-32 dataset retrieval experiments, including baseline network and pre-trained on LogoDet-3K network. We used the latest retrieval based detection method Deepvision~\cite{II2019Scalable}, which adopts two different state-of-the-art classification architectures, namely ResNet101 and DenseNet161, and the experimental results are 52.62\% and 50.38\%, respectively. The pre-train model on LogoDet-3K is used to the baseline method Deepvision~\cite{II2019Scalable}, the results are 54.17\% and 52.91\% mAP, with the 1.55\% and 2.31\% improvement compared with Deepvision. The experimental results show that the pre-trained model generated by our dataset is also effective in the logo retrieval task, further illustrating the value of LogoDet-3K in logo-related research.

\begin{table}[!t]
	\caption{Evaluation retrieval results on FlickrLogos-32 (\%).}
	\label{Retrieval}
	\centering
	\setlength{\tabcolsep}{1.8mm}{
		\begin{tabular}{cc}
			\hline
			Method                                                     & mAP \\ \hline
			baseline~\cite{Su2018Deep}                                 & 36.00  \\ 
			ResNet101                                                  & 32.70   \\
			DenseNet161                                                & 36.80  \\
			ResNet101+Litw~\cite{Andras2017Open}                       & 46.40  \\
			DenseNet161+Litw~\cite{Andras2017Open}                     & 44.80  \\
			Deepvision(ResNet101)                                      & 52.62\\
			Deepvision(DenseNet161)                                    &50.78\\
			\textbf{Deepvision(ResNet101+Pre-trained)}                 &\textbf{54.17}   \\
			\textbf{Deepvision(DenseNet161+Pre-trained)}               &\textbf{52.91}   \\
			
			\hline
	\end{tabular}}
\end{table}
\begin{figure*}[!htbp]
	\centering
	\includegraphics[width=1\textwidth]{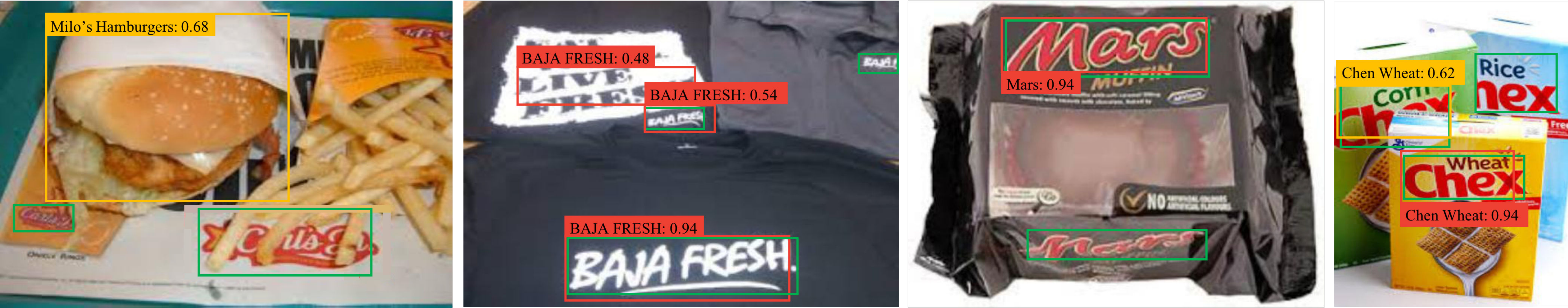}
	\caption{Qualitative result of some failure cases on Logo-Yolo. Green boxes denotes the ground-truth. Red boxes represent correct logo detections, while yellow are mistakes.}
	\label{wrong}
\end{figure*}
\subsection{Discussion}
Compared with existing methods, our proposed method obtains better detection performance, especially in solving small objects and complex backgrounds of logo images compared with YOLOv3. However, it can not achieve high detection performance for some cases. Fig.~\ref{wrong} shows some failure cases from Logo-Yolo. Logo-Yolo is difficult to detect the smaller scale logos, leading to missed detection, such as the third image in Fig.~\ref{wrong}. In addition, the logos under the same brand are similar and often appear in the same image, so there will be some problems in the object classification, such as the four images. As shown in Fig.~\ref{wrong}, we found that our method encountered lower performance when the following cases appear, such as blocked logo objects, logo objects close to the background and very small objects. Therefore, the logo detection on LogoDet-3K still has great challenges, such as the multi-label problem and large-scale problem, and it meanwhile highlights the comparative difficulty of the LogoDet-3K dataset.

\section{Conclusions}
In this paper, we present LogoDet-3K dataset, the largest  logo detection dataset with full annotation, which has 3,000 logo categories, about 200,000 high-quality manually annotated logo objects and 158,652 images. Detailed analysis shows the LogoDet-3K was highly diverse and more challenging than previous logo datasets. Therefore, it establishes a more challenging benchmark and can benefit many existing localization sensitive logo-relate tasks. In addition, we propose a new strong baseline method Logo-Yolo, which can get better detection performance than other state-of-art baselines. And we also report results of various detection models and demonstrate the effectiveness of our method and better generalization ability on other three logo datasets and logo retrieval tasks. 

In the future, we hope LogoDet-3K will become a new benchmark dataset for a broad range of logo related research. Such as logo detection, logo retrieval and logo synthesis tasks. With the rapid development of major brands, real-time logo detection will become the trend of future research. We will continue to explore the characteristics of the LogoDet-3K dataset, and use anchor-free and lightweight design methods specifically for logo detection to achieve faster and more accurate logo detection.
\bibliographystyle{IEEEtran}
\bibliography{Logo_detection}  

\end{document}